%% file: acl_latex.tex
\pdfoutput=1

\documentclass[11pt]{article}

\usepackage{acl}  

\usepackage{colortbl}

\usepackage{times}
\usepackage{latexsym}

\usepackage[T1]{fontenc}

\usepackage[utf8]{inputenc}

\usepackage{microtype}

\usepackage{graphicx}
\usepackage[group-separator={,}]{siunitx}
\usepackage{hyperref}
\usepackage{multirow,tabularx}
    \newcolumntype{L}{>{\raggedright\arraybackslash}X}
    \newcolumntype{V}{>{\raggedleft\arraybackslash}m{1.5cm}}
\hypersetup{
colorlinks   = true,
citecolor    = blue,
urlcolor    =  blue
}
\usepackage{listings}
\usepackage{framed}
\usepackage{amsmath}
\usepackage{hhline}
\usepackage{caption}
\usepackage{subcaption}
\input{footnotes}
\usepackage{cuted}
\title{Synthesis and Evaluation of a Domain-specific Large Data Set for Dungeons \& Dragons}

\author{Akila Peiris \and Nisansa de Silva \\
  Department of Computer Science \& Engineering,\\University of Moratuwa, Sri Lanka \\
  \texttt{\{akila.21,nisansadds\}@cse.mrt.ac.lk} \\}

\newcommand{\FR}{\textit{FRW}}
\newcommand{\FRW}{\FR{} data set}
\newcommand{\FRWP}{\FR\textit{-P}}
\newcommand{\FRWJ}{\FR\textit{-J}}
\newcommand{\FRWFJ}{\FR\textit{-FJ}}
\newcommand{\FRWL}{\FR\textit{-L}}
\newcommand{\FRWFL}{\FR\textit{-FL}}
\newcommand{\FRWCL}{\FR\textit{-CL}}
\newcommand{\FRWI}{\FR\textit{-I}}
\newcommand{\FRWPE}{\FR\textit{-PE}}
\newcommand{\FRWW}{\FR\textit{-W}}
\newcommand{\FRWD}{\FR\textit{-D}}
\newcommand{\FRWFD}{\FR\textit{-FD}}
\newcommand{\FRWTGen}{Forgotten Realms Free Text Generator}

\newcommand{\dnd}{D\&D}
\newcommand{\keyvalue}{key-value}

\newcommand{\articles}{over 45,200 articles}
\newcommand{\articleDate}{September 2022}

\begin{document}
\maketitle
\begin{abstract}

This paper introduces the Forgotten Realms Wiki (\FR) data set and domain specific natural language generation using \FR{} along with related analyses. Forgotten Realms is the de-facto default setting of the popular open ended tabletop fantasy role playing game, Dungeons \& Dragons. The data set was extracted from the Forgotten Realms Fandom wiki consisting of more than \articles. The \FRW{} is constituted of 11 sub-data sets in a number of formats: raw plain text, plain text annotated by article title, directed link graphs, wiki info-boxes annotated by the wiki article title, Poincaré embedding of first link graph, multiple Word2Vec and Doc2Vec models of the corpus. 
This is the first data set of this size for the Dungeons \& Dragons domain. We then present a pairwise similarity comparison benchmark which utilizes similarity measures.
In addition, we perform \dnd{} domain specific natural language generation using the corpus and evaluate the named entity classification with respect to the lore of Forgotten Realms.

\end{abstract}


\section{Introduction}
Specialized and domain specific data sets are useful for a number of advanced tasks in the domain of Natural Language Processing (NLP). For example, recent studies have shown that the domain specificity significantly impacts vital NLP tasks such as measuring semantic similarity~\cite{sugathadasa2017synergistic} and domain specific text generation~\cite{lebret-etal-2016-neural}. Further, it has been shown that models developed using data from a generic domain do not seamlessly transfer to tasks in a specific domain~\cite{rajapaksha2020rule}. 
Fantasy domains could be considered an extreme case of domain specific data, as it is possible to observe the full spectrum of deviations from the non-domain specific (general domain) usage, both in the lexical and semantic perspectives. An example for lexical differences is the usage of \textit{dwarves} as the plural form of \textit{dwarf} in the fantasy genre\footnote{This is inherited from the spelling used in the \textit{Lord of the Rings} and other relevant publications by J. R. R. Tolkien} in place of the general domain spelling \textit{dwarfs}. An example for semantic differences can be seen in the words \textit{Ghost}\footURL{https://bit.ly/DnDGhost} and \textit{Wraith}\footURL{https://bit.ly/DnDwraith}. In the Merriam-Webster dictionary, they are given as synonyms in the generic domain\footURL{https://www.merriam-webster.com/thesaurus/ghost} while in the domain of the fantasy role playing game Dungeons \& Dragons, they are defined as two distinct creatures. In this paper, we present the \FRW{}, specific to the \textit{Forgotten Realms} setting of \textit{Dungeons \& Dragons}. We expect our data set to be useful for in-domain tasks such as text generation~\cite{zhang2019bertscore}, information extraction~\cite{de-silva-dou-2021-semantic}, and information retrieval~\cite{sugathadasa2018legal}. We also anticipate our data set being vital for cross domain tasks such as text alignment~\cite{sanchez2014winning}, style transfer~\cite{fu2018style}, and summarizing~\cite{el2020automatic}. As a primer for these usages, we introduce a pairwise similarity comparison benchmark and evaluate the domain-specific free text generation task.

\subsection{Dungeons \& Dragons}

Dungeons \& Dragons (\dnd{} or DnD), is an open-ended pen and paper tabletop role playing game (RPG) which has been commercially available since \citet{gygax1974dungeons} published the first version.
The games are primarily based on fantasy genre. However, there is a plethora of other settings ranging from science fiction, post-apocalyptic to hollow world and much more. Even within a selected genre, it is highly customizable, for example, a fantasy setting might be in high or low fantasy. 
D\&D has a predefined set of rules governing almost every aspect of the gameplay including the \textit{setting}. A setting has a lore, species and artifacts among other components; which can be dissimilar between settings. There are also several editions of D\&D, with 5~\cite{crawford2014player} being the latest. It is the version that our \FRW{}s are predominantly based on. However, it does contain some information from earlier editions in cases where there have been changes to the lore between versions or in cases where information have been consistently brought forward.

\subsection{Forgotten Realms Wiki}
Forgotten Realms as mentioned, is a setting which is categorized under \textit{high fantasy}, set in an alternate world filled with magical elements combined with larger than life themes, plots, and characters. 
It originated as a medieval European setting but over the years, has been influenced by other cultures including Middle Eastern and Asian. \textit{Forgotten Realms} became the most utilized of all the official \dnd{} settings after it became the de-facto default setting of the immensely popular~\cite{whitten_dungeons_2021} 5th edition. 
Almost all of the official material published for D\&D is based on this setting. 
Due to this, \textit{Forgotten Realms} now has the most resources and information available from all the settings in \dnd. 

However, this massive amount of information is distributed among hundreds of official books and magazines making it intractable for a casual enthusiast of \dnd. To remedy this problem and to curate and consolidate the information, the community of \dnd{} enthusiasts voluntarily contribute and maintain the \textit{Forgotten Realms wikia}~\footURL{https://forgottenrealms.fandom.com}.
A \textit{Wikia} or a \textit{Fandom Wiki} is a Wikipedia~\footURL{https://en.wikipedia.org}-esque website (uses the same MediaWiki~\footURL{https://www.mediawiki.org/wiki/MediaWiki} collaborative documentation platform) hosted by Fandom, Inc.~\footURL{https://www.fandom.com}. This is typically dedicated to a particular domain (e.g., Star Wars~\footURL{https://starwars.fandom.com/}, Marvel~\footURL{https://marvel.fandom.com/},Harry Potter~\footURL{https://harrypotter.fandom.com/}, Formula One Racing~\footURL{https://f1.fandom.com/wiki/}). 
The \textit{Forgotten Realms Wikia} has \articles\footnote{This is the statistic shown in the website but there are other articles that you can export which are not considered for this parameter} as of \articleDate{} and keeps growing at a rapid pace. 

\subsection{Wikipedia and other Wikis as Data Sources}

Wikipedia and other Wikia, maintained by a community of volunteers, are treasure troves of domain specific knowledge~\cite{ferrari2017detecting}. While there are endless debates regarding the validity of such community maintained knowledge-bases in scientific context~\cite{cozza2016matter,ferschke2014quality}, there are still a number of ways they can be used to further the scientific frontiers~\cite{zesch2007analyzing,ponzetto2007knowledge,zesch-gurevych-2007-analysis,zesch-etal-2008-extracting}. One such usage in the field of Natural Language Processing is to use them as data sources which not only provide corpora of the relevant domains but also provides insight into community-based collaborative text maintenance~\cite{ferschke2013survey}. 

\begin{figure}[!ht]
     \centering
     \begin{subfigure}[b]{0.23\textwidth}
         \centering
         \includegraphics[width=\textwidth]{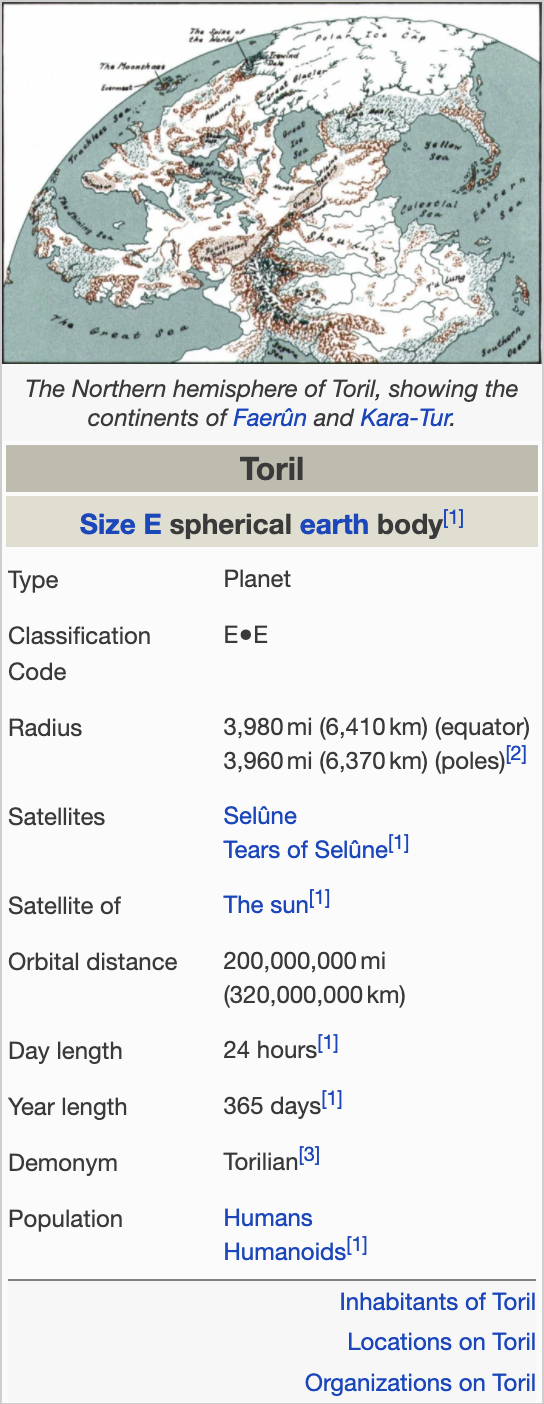}
         \caption{In-game location}
         \label{fig:toril}
     \end{subfigure}
     \hfill
     \begin{subfigure}[b]{0.23\textwidth}
         \centering
         \includegraphics[width=\textwidth]{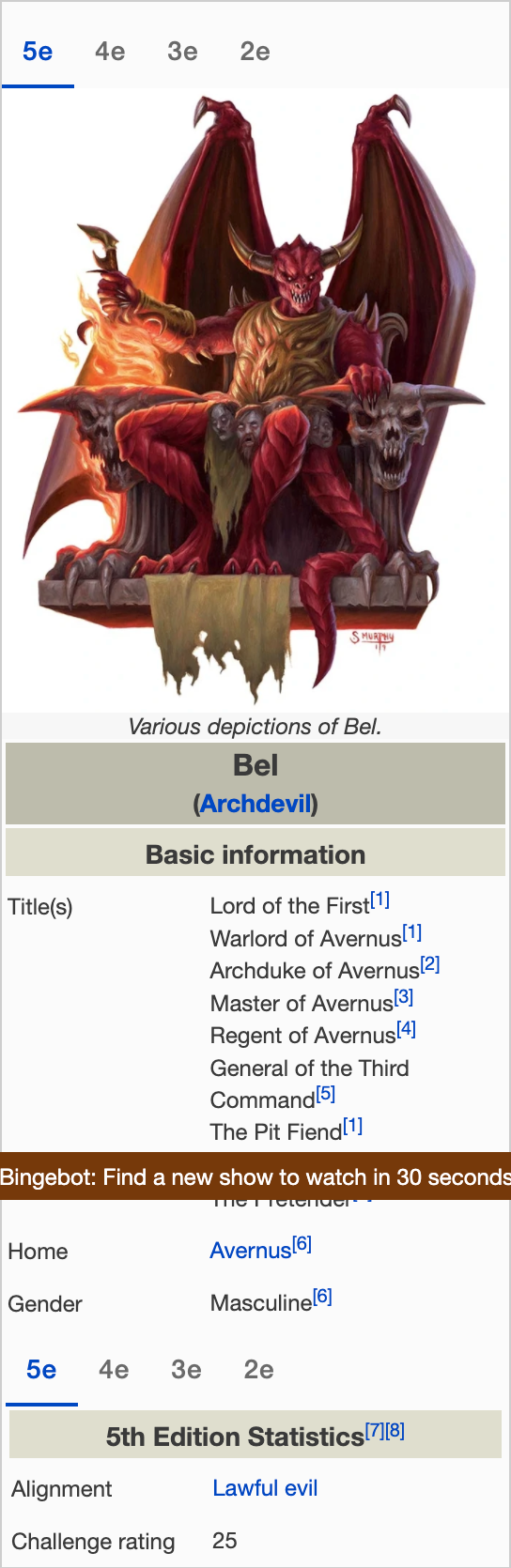}
         \caption{Infobox with multiple editions}
         \label{fig:bel}
     \end{subfigure}
     \hfill
     \begin{subfigure}[b]{0.23\textwidth}
         \centering
         \includegraphics[width=\textwidth]{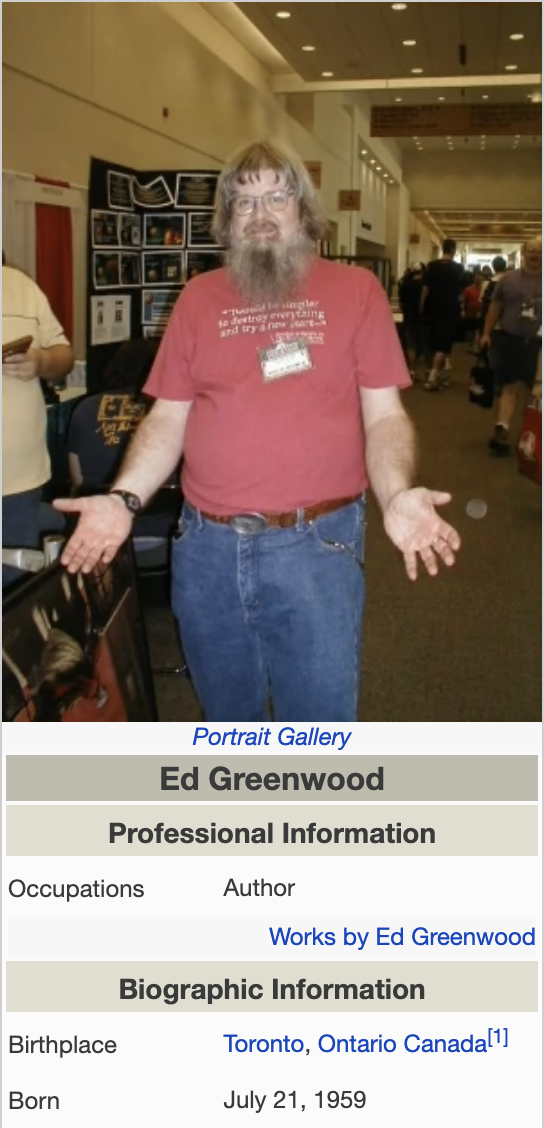}
         \caption{Real-world\\author}
         \label{fig:ed}
     \end{subfigure}
     \hfill
     \begin{subfigure}[b]{0.23\textwidth}
         \centering
         \includegraphics[width=\textwidth]{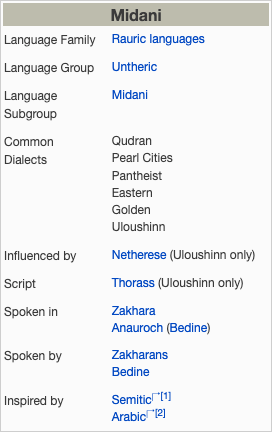}
         \caption{Infobox without an image}
         \label{fig:karsuss}
     \end{subfigure}
\caption{Some of the different types of infoboxes in \FR}  
\label{infobox}
\end{figure}

The possibility of accessing as a freely available source in multiple languages~\cite{nastase2013transforming} (human translated), being extensive, and having special information content such as infoboxes\footURL{https://en.wikipedia.org/wiki/Help:Infobox} make Wikipedia and similar wikia rich resources for data. An infobox is the table-like structure typically found at the top-right side of a wiki article (Fig.~\ref{infobox}). It is a  human annotated, tabular summary of the article, arranged in a \keyvalue{} structure according to a template. According to~\citet{lange2010extracting} about one third of all Wikipedia articles contain an infobox. While this is indeed a rich source of information, they are known to be noisy and sparse\cite{hoffmann-etal-2010-learning}. The wiki page itself only renders the pairs that contain values. 

Another special information content is found in the first paragraph (lead section\footURL{https://en.wikipedia.org/wiki/Wikipedia:Manual_of_Style/Lead_section}) of a wiki article. According to the guideline, this is typically formatted as an abstractive summary to the entire page. In their study on wikipedia,~\citet{lange2010extracting} report that there is a 92\% chance to find any of the information summarized in the infobox within the first paragraph.

\subsection{Domain Specific Text Generation}
Domain specific text generation is an emerging area in NLP~\cite{liu2018table,chen-etal-2021-scixgen-scientific,zhang2022survey,amin-nejad-etal-2020-exploring}. The objective in this is to generate text which adheres to a given domain, in the sense that the content generated should be semantically and pragmatically truthful to the said domain. One of the reasons why domain specific text generation is difficult compared to generic text generation is that, in most cases this requires copious amounts of linguistic resources based on the domain in question. This hurdle is true even for fine-tuning a pre-trained model which relatively demands less amount of data than training a model from ground-up~\cite{zhang2021dsgpt}.

\section{Related Work}

\subsection{Wikipedia and Other Wiki Data Sets}
In recent times the availability of linguistic data sources have increased significantly. Especially Wikipedia based data sets such as Wit~\cite{srinivasan2021wit}, WCEP~\cite{gholipour-ghalandari-etal-2020-large}, and SQuAD~\cite{rajpurkar-etal-2016-squad}. 
Tools such as \textsc{LUCHS}~\cite{hoffmann-etal-2010-learning} and \textsc{WOE}~\cite{wu-weld-2010-open} are capable of extracting information from Wikipedia pages to create such data sets. Both systems rely on the \keyvalue{} structure of the infoboxes to guide the information extraction process from the natural language text. This guided process is akin to the widely used Ontology-Based Information Extraction (OBIE)~\cite{de2017discovering}. 

As mentioned, Fandom, Inc. is an organization which hosts wikis for a large number of entertainment media franchises and other areas as the general populace may desire. The Fandom wikis operate on the same technology and guidelines\footURL{https://www.mediawiki.org/} as Wikipedia. They are good sources of domain specific data for different media franchises as they are written in the desired domain and offers a clear demarcation from in-domain and out-domain data as opposed to obtaining data from sources such as the common crawl~\cite{kreutzer-etal-2022-quality}. 
The Critical Role Dungeons \& Dragons Data set (\textit{CRD3})~\cite{rameshkumar-bailey-2020-storytelling} is a \dnd{} domain-specific, narrative driven, multi speaker dialog  data set that has been extracted from a similar Fandom Wiki\footURL{https://criticalrole.fandom.com}. This particular wiki is dedicated to the web series, \textit{Critical Role}, a live \dnd{} gameplay series. The data set consists of multi-speaker dialogue that form a narrative, paired with their abstractive summaries.

\subsection{Domain Specific Text Generation}
Text generation methodologies fall into three categories~\cite{stent-etal-2004-trainable}. \textbf{Template based} methods~\cite{busemann-horacek-1998-flexible,reiter1997building,mcroy2003augmented} are the most common variant. It uses pre-defined text templates applicable to different scenarios to generate text. It is a tedious and non-salable approach. 
Secondly, there is \textbf{Rule based} generation~\cite{dale2003coral,turner-etal-2009-generating,reiter2005choosing}. This has three inter-dependent phases: (1) text planning - governs the process of meaning representation retrieval from a knowledge base, (2) sentence planning - governs the words and their order to produce coherent sentences, and (3) surface realization - converts the sentence plan into actual sentences. 
Thirdly, the \textbf{Data driven} approach~\cite{barzilay-lapata-2005-collective,liang-etal-2009-learning}. Unlike rule based approaches, data driven ones require more data. This burden is alleviated using pre-trained language models and transfer learning techniques. Open AI's Generative Pre-trained Transformer models GPT-2~\cite{radford2019language}, GPT-3~\cite{brown2020language} as well as the open source reproductions of these, the GPT-Neo models~\cite{gao2020pile,black-etal-2022-gpt} are such large pre-trained language models. In addition to having been trained on very large data sets, they are also large networks. These models are capable of generating highly sophisticated texts. With some fine-tuning, they can be adapted to do the same for specific domains.


\section{Forgotten Realms Wiki (\FR) Data set}
We introduce the Forgotten Realms Wiki (\FR) data set\footURL{https://huggingface.co/Akila/ForgottenRealmsFreeTextGenerator}, extracted from the Forgotten Realms Fandom wikia. We have extracted multiple data sets from this textual resource and present them in multiple formats for different linguistic use cases. We also present several different embeddings for this data set including Poincaré hierarchical embedding and multiple word and document level embeddings. A summary of the data sets is shown in Table~\ref{table:dataset} and the individual statistics for each data set can be found listed under Table~\ref{table:statistics}.

\begin{figure*}[!htb]
\begin{minipage}[t]{0.48\textwidth}
    \begin{subfigure}[!hbt]{0.48\textwidth}
         \centering
          \includegraphics[width=\textwidth]{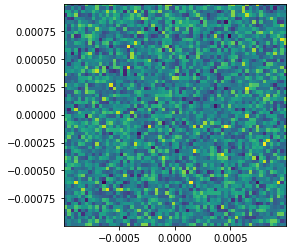}
          \caption{Initial (0 epochs)}
     \end{subfigure}
     \begin{subfigure}[!hbt]{0.48\textwidth}
          \centering         \includegraphics[width=\textwidth]{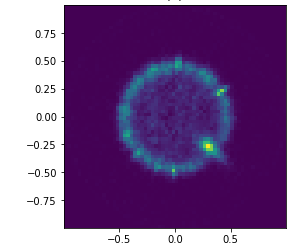}
          \caption{After 50 epochs}
     \end{subfigure}
     
     \caption{Poincaré Embedding convergence}
     \label{fig.poincare}
\end{minipage}%
\hfill
\begin{minipage}[t]{0.45\textwidth}
\begin{subfigure}[!hbt]{\textwidth}
\centering
\includegraphics[width=\textwidth]{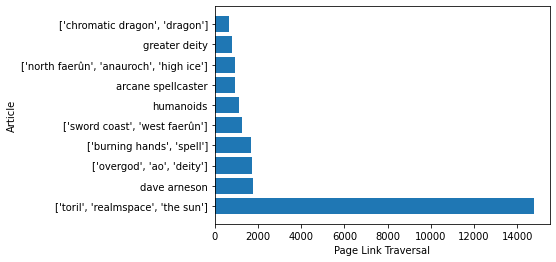}
\end{subfigure}
\caption{First link traversal graph}
\label{fig.first_link}
\end{minipage}
\end{figure*}

The plain text corpora (\FRWP, \FRWJ) are devoid of special data structures and other markings. As for the links, the MediaWiki allows having an alternative text to display for the links instead of the exact page title for aesthetics of the writing, hence we extract that part for the plain text corpus when available. 
The \FRWFJ{} data set is composed of mainly the \textit{lead sections}\footURL{https://en.wikipedia.org/wiki/Wikipedia:Manual_of_Style/Lead_section}. Because of this, we can consider this as an abstractive summary of \FRWJ. 
The \FRWCL{} links pages with categories. The categories themselves have rendered pages which aggregate the pages under each category.
The infobox data in \FRWI{} are converted from markdown to JSON before being embedded in the overall JSON structure indexed by the page title.
Each of the word and document embedding data sets (\FRWW, \FRWD, \FRWFD) have 2 different embeddings used. For word level embeddings CBOW and Skip-gram~\cite{mikolov2013efficient} are used. While PV-DBOW and PV-DM~\cite{le2014distributed} are used in document embeddings. Figure~\ref{fig.poincare} shows the convergence of the Poincaré embedding data set, \FRWPE.

All of the data sets that use a JSON structure (\FRWJ, \FRWFJ, \FRWI) use the same high level JSON schema. The pages are organized in a JSON array with \textit{page} and \textit{content} being the only two attributes in each element. The \textit{page} attribute contains the article title while the \textit{content} attribute contains the corresponding information extracted from the page. This information is in plain text format for \FRWJ{} and \FRWFJ. In the case of \FRWI, it is a JSON dictionary containing the infobox content as the key-value pairs. Code~\ref{JSONexample} shows the top level JSON schema.

\begin{lstlisting}[caption={JSON top level structure},label=JSONexample]
[
  ...,
  {
      "page": "page_title",
      "content": "page_content"
  },
  ...
]
\end{lstlisting}

\begin{table}[!h]
\begin{center}
\begin{tabularx}{\columnwidth}{|l|X|}
      \hline
      \textbf{Name}&\textbf{Description}\\
      \hline
      \FRWP & Raw plain text corpus (no Markdown text markings)\\
      \hline
      \FRWJ & A JSON structure with plain text indexed by article title\\
      \hline
      \FRWFJ & A JSON structure with only the first paragraph (plain text) of the articles indexed by article title\\
      \hline
      \FRWL & A directed graph indicating all the references in the articles to other articles \\
      \hline
      \FRWFL & A directed graph indicating the first references in the articles to other articles \\
      \hline
      \FRWCL & A directed graph indicating the category references in the articles to category pages.\\
      \hline
      \FRWI & A JSON structure for the Wikipedia infobox substructures indexed by article title\\
      \hline
      \FRWPE & Poincaré embedding of \FRWFL\\
      \hline
      \FRWW & 2 Word2Vec models for \FRWP{} (CBOW and Skip-gram)\\
      \hline
      \FRWD & 2 Doc2Vec models for \FRWP{} (PV-DBOW and PV-DM)\\
      \hline
      \FRWFD & 2 Doc2Vec models for \FRWFJ{} (PV-DBOW and PV-DM)\\
      \hline
\end{tabularx}
\caption{\FRW}
\label{table:dataset}
\end{center}
\end{table}


\begin{table*}[!htb]

    \begin{subtable}[!htb]{0.48\textwidth}
    \begin{center}
    \begin{tabularx}{\columnwidth}{|L|V|}
          \hline
          \textbf{Statistic} & \textbf{Value} \\
          \hline
          Total number of tokens (excluding article titles) & \num[group-separator={,}]{9189536}\\
          \hline
          Total number of tokens (including article titles) & \num[group-separator={,}]{9287670}\\
          \hline
          Total number of unique tokens & \num[group-separator={,}]{145624}\\
          \hline
          Total number of sentences & \num[group-separator={,}]{517248}\\
          \hline
    \end{tabularx}
    \end{center}
    \caption{$\FRWP$}
    \label{tab:FRP}
    \end{subtable}
    \hfil
    \begin{subtable}[!htb]{0.48\textwidth}
    \begin{center}
    \begin{tabularx}{\columnwidth}{|L|V|}
          \hline
          \textbf{Statistic} & \textbf{Value} \\
          \hline
          Total number of articles & \num[group-separator={,}]{48892}\\
          \hline
          Average number of tokens  per sentence & \num[group-separator={,}]{ 17.77}\\
          \hline
          Average number of tokens per article & \num[group-separator={,}]{187.96}\\
          \hline
          Average number of sentences  per article & \num[group-separator={,}]{ 10.58}\\
          \hline
    \end{tabularx}
    \end{center}
    \caption{$\FRWJ$}
    \label{tab:FRJ}
    \end{subtable}
    \hfill
    \begin{subtable}[!htb]{0.48\textwidth}
    \begin{center}
    \begin{tabularx}{\columnwidth}{|L|V|}
          \hline
          \textbf{Statistic} & \textbf{Value} \\
          \hline
          Total number of articles & \num[group-separator={,}]{41204}\\
          \hline
          Total number of tokens & \num[group-separator={,}]{980047}\\
          \hline
          Total number of sentences & \num[group-separator={,}]{98244}\\
          \hline
          Average number of tokens  per sentence & \num[group-separator={,}]{ 9.98}\\
          \hline
          Average number of tokens per article & \num[group-separator={,}]{23.78}\\
          \hline
          Average number of sentences  per article & \num[group-separator={,}]{2.38}\\
          \hline
    \end{tabularx}
    \end{center}
    \caption{$\FRWFJ$}
    \label{tab:FRFJ}
    \end{subtable}
    \hfill
    \begin{subtable}[!htb]{0.48\textwidth}
    \begin{center}
    \begin{tabularx}{\columnwidth}{|L|V|}
          \hline
          \textbf{Statistic} & \textbf{Value} \\
          \hline
          Average number of attributes per infobox & \num[group-separator={,}]{40.54}\\
          \hline
          Average number of completed (filled) attributes per infobox & \num[group-separator={,}]{10.40}\\
          \hline
          Total number of articles containing infoboxes & \num[group-separator={,}]{35923}\\
          \hline
    \end{tabularx}
    \end{center}
    \caption{$\FRWI$}
    \label{tab:FRI}
    \end{subtable}
    \hfill
    \begin{subtable}[!htb]{0.48\textwidth}
    \begin{center}
    \begin{tabularx}{\columnwidth}{|L|V|}
          \hline
          \textbf{Statistic} & \textbf{Value} \\
          \hline
          Total number of nodes & \num[group-separator={,}]{46910}\\
          \hline
          Total number of edges & \num[group-separator={,}]{570857}\\
          \hline
          Average number of edges per node & \num[group-separator={,}]{12.16}\\
          \hline
    \end{tabularx}
    \end{center}
    \caption{$\FRWL$}
    \label{tab:FRL}
    \end{subtable}
    \hfill
    \begin{subtable}[!htb]{0.48\textwidth}
    \begin{center}
    \begin{tabularx}{\columnwidth}{|L|V|}
          \hline
          \textbf{Statistic} & \textbf{Value} \\
          \hline
          Total number of nodes & \num[group-separator={,}]{43329}\\
          \hline
          Total number of edges & \num[group-separator={,}]{41213}\\
          \hline
          Number of nodes not referenced by others & \num[group-separator={,}]{34881}\\
          \hline
          Number of nodes with no references & \num[group-separator={,}]{2151}\\
          \hline
    \end{tabularx}
    \end{center}
    \caption{$\FRWFL$}
    \label{tab:FRFL}
    \end{subtable}
\caption{Statistics of different sub data sets of the $\FRW$}
\label{table:statistics}
\end{table*}

 \newcommand{\twoMulti}{0.23}

\section{Use Case Analysis 1: Semantic Similarity Comparison}

To illustrate both the consistency as well as the non-trivial nature of data sets we have collected, we have performed similarity calculations for a set of text pairs extracted from the data set using multiple different similarity metrics. By the high alignment of semantic similarity in similar perspective data sets, we show the consistency in our data sets. By the low alignment of the differing perspective data sets, we show that the individual data sets are not redundant and that they carry unique information that may not have overlaps with other data sets that we present in this work.

\subsection{Text Pairs for Evaluation}
To ensure that these different metrics are comparable, we have used the same set of text pairs for all of the similarity calculations. Hierarchical similarities are measured using article titles and the embedded vector distance based similarities are calculated using article contents. We use the \FRWFJ{} data set to generate the text pairs. Since the \FRWFJ{} data set is a subset of \FRWJ, it ensures that a) all the pairs correspond to actual wiki articles b) has valid text content c) full document and first paragraph only data sets are available for different similarity calculations, and d) almost all nodes (page titles) are significant/``worthy of notice''\footURL{https://en.wikipedia.org/wiki/Wikipedia:Notability} to the domain as per the Wikipedia guidelines.

\subsubsection{First Link} 
\label{subsec:first_link}
First link is the first internal reference link (refers to another article in the same wikia) found in an article that is not a broken link or a miscellaneous link such as the pronunciation guide. According to the guidelines\footURL{https://en.wikipedia.org/wiki/Wikipedia:Manual_of_Style/Lead_section}, the lead section of a typical Wikipedia article contains links to other articles that provide context to the article in question i.e., the references in lead section point towards more generalized concepts and/or any other concepts related to the context of that article. We use this arrangement to measure the similarity or relatedness of topics. 
This leads to an interesting pattern where clicking the the first link of a random Wikipedia page and doing so repeatedly on the subsequent pages will 97\% of the time~\cite{lamprecht2016evaluating} lead to the article ``Philosophy''\footURL{https://en.wikipedia.org/wiki/Philosophy} which forms a cycle with the article ``Metaphysics''\footURL{https://en.wikipedia.org/wiki/Metaphysics}. The rest of these \textit{first link traversals} exhibit one of the following shortfalls: 1) contain no internal links, 2) contains a self loop, 3) ends up in an isolated tree, 4) form a cycle with a few other pages.
The Forgotten Realms wiki also abide by the same principle. The \textit{center} of the Forgotten Realms wiki universe is a cycle composed of the articles, ``\textit{Toril}'', ``\textit{Realmspace}'' and ``\textit{The Sun}''. However, unlike in the case of Wikipedia, in Forgotten Realms wiki, this only applies to around 30.2\% of the articles. Figure~\ref{fig.first_link} lists the 10 most commonly traversed articles using this method. The ones enclosed in ``[ ]'' refer to cycles, for example ['\textit{toril}', '\textit{realmsphere}', '\textit{the sun}'] refers to a first link cycle between the three corresponding articles.

\subsubsection{Issues with Category Links for Semantic Similarity Evaluation}
Even though it is the de-facto categorical hierarchy, there are many issues with using category links as a measure for semantic similarity. The most prominent bottleneck of \FRWCL{} data set is that it is mostly a flat hierarchy. So any set of node pairs would have almost identical distance measures no matter how different they are semantically. Secondly, the Categories are not consistent across all articles, i.e,. while some articles may have an abundance of Categories, others may have have little to none. Finally, Category pages do not necessarily have article content as a typical page does, hindering the ability to perform effective word and document embedding. 

\subsubsection{Generating Text Pairs for Evaluation} 
We created 1,000,000 unique text pairs using 41,000 nodes from \FRWFL. We have also ensured that there are no interchangeable duplicates. To ensure that the selected pairs have better representation, we have used a weighted random sampling technique with dynamically updated weights. The sampling was done with replacement. The probability of an item $i$ getting selected for the sample pair set is given in Equation~\ref{eq:pair_prob}, where $N$ is the total number of pairs we generate and $k_j$ is the number of times the $j$th item has already been selected.






\subsection{Hierarchical Similarity Measures}

We have used the \FRWFL{} data set as the base for similarity measures using hierarchical similarity evaluation methods. Although the \FRWFL{} is already devoid of any self-loops, there are cycles and isolated trees while also lacking a common root node. We process this and convert into a directed cyclic graph.

Let $G$ be a graph in the set of disconnected graphs $G = (V, E) \in {G_1, G_2,..., G_n} $ where $E \in E^\prime $ and $ V \in V^\prime$. $G_c$ represent a subgraph of a given graph $G_c = (V_c, E_c) \in {G} $ where $e_1, . . . , e_n$ is a trail with vertex sequence $a_1, . . . , a_n$ (cyclic graph). Then $\forall G \in {G_1,...,G_n}$ apply Equation~\ref{eq:graph} to obtain the final unified graph, $G' = (V^\prime, E^\prime)$.




For the intermediate node $v^\prime$, we use a comma separated combination of the names of the nodes forming the cycle. Using an intermediate node helps us retain the relatedness they had with one another up to a certain degree before reaching the root node. We use algorithmic measures such as~\citet{wu-palmer-1994-verb} similarity metric and~\citet{jiang-conrath-1997-semantic} distance measure, both of which use the Least Common Ancestor ($LCA$) as the basis for the calculations. Apart from this, we have also evaluated with the hierarchical embedding using Poincaré~\cite{nickel2017poincare} method. While other embedding methods such as ones using metadata~\cite{xing-paul-2017-incorporating, zhou-etal-2015-learning} can be experimented with, we have chosen the Poincaré embedding since we are measuring hierarchical similarities. This allows us increase the comparability between the different types of measurements in our experiment. 

It should be noted that, for the sake of this comparative analysis, we have converted the Jiang-Conrath distance measure~\cite{jiang-conrath-1997-semantic} into a similarity measure ranging from $0$ to $1$ as shown in the Equation \ref{eq:jiang} where the $LCA(a,b)$ function returns the least common ancestor of the nodes $a$ and $b$, the $IC(d)$ function returns the Information Content of the node $d$, and $c_i$ is the node in the hierarchy representing the term $t_i$. 


\subsection{Embedding Based Similarity Measures}
We have performed both word embedding on \FRWP{} corpus and document embedding for the \FRWJ{} data set to create \FRWW{} and \FRWD{} data sets. Both of these data sets contain essentially the same content albeit the format. In addition, we have created \FRWFD{} data set using \FRWFJ{} which only contains the first paragraph of each page to evaluate the effectiveness of the first paragraph in comparison to the whole text. For embedding vector distance based similarity, we have used the \FRWW{} data set containing CBOW and Skip-gram~\cite{mikolov2013efficient} model embeddings. For word embedding, when a title is given the corresponding article is retrieved from \FRWJ{}. Then for all the words in the article, the word vectors are fetched from the \FRWW{} data set and a single vector is obtained via average pooling. Cosine similarity is defined as shown in Equation~\ref{eq:cosine}, where $t_i$ is a string in the domain and $v_i$ is the corresponding vector in the embedded vector space. 

\begin{strip}
\begin{align}
    P(i) = \frac{\sqrt{N} - k_{i}}{N\sqrt{N}-\sum_{j=1}^{N}k_{j}} 
    \label{eq:pair_prob}
\end{align}

\begin{align}
    E^\prime  = 
    \begin{cases}
        E \cup (root, v)                               & \text{if } deg^-(v) = 0  \text{ and } v\neq root\\
        E \cup (root, v^\prime) \cup (v^\prime, v)     & \text{if } v \in G_c  \\
        E                                              & \text{otherwise}
    \end{cases}
    \label{eq:graph}
\end{align}

\begin{align}
    JC \_ sim(t_1,t_2) = \frac{1}{1 + \lvert 2 \times IC(LCA(c_1,c_2)) - ( IC(c_1) + IC(c_2) ) \rvert}
    \label{eq:jiang}
\end{align}

\begin{align}
    cosine \_ similarity (t_1,t_2) = \frac{v_1 \cdot v_2}{\lvert\lvert v_1 \rvert\rvert\lvert\lvert v_2 \rvert\rvert}
    \label{eq:cosine}
\end{align}

\end{strip}

For Doc2vec~\cite{le2014distributed}, we have used both \FRWD{} and \FRWFD{} data sets with each having a \texttt{PV-DBOW} and a \texttt{PV-DM} model~\cite{le2014distributed}  resulting in four embedded models altogether. The four models are trained with the article title as the tag for the content of the article. Hence the document vector itself can be fetched directly from the model using the article title (text phrase).

\newcommand{\Gray}{
 \cellcolor[gray]{0.8}
}

\newcommand{\Number}[1]{
 {\small #1}
}

\begin{table*}[!htb]
\begin{center}
\begin{tabular}{|c|c|c|c|c|c|c|c|c|c|c|c|c|}

      \hline
   \multicolumn{4}{|c|}{}&    \multicolumn{3}{c|}{\textbf{Hierarchical}}  
        &\multicolumn{6}{c|}{\textbf{Embedding}}\\
      \hhline{~~~~---------}
   \multicolumn{4}{|c|}{}& \multirow{2}{*}{\multirow{2}{*}{\textbf{WP}}} &
      \multirow{2}{*}{\multirow{2}{*}{\textbf{JC}}} &
      \multirow{2}{*}{\multirow{2}{*}{\textbf{P}}} & 
      \multicolumn{2}{c|}{\textbf{Word2Vec}} & 
      \multicolumn{4}{c|}{\textbf{Doc2Vec}}  \\
      \hhline{~~~~~~~------}
   \multicolumn{4}{|c|}{}&  &  &  &  \multicolumn{2}{c|}{(\FRWJ)}  &  \multicolumn{2}{c|}{(\FRWFJ)} & \multicolumn{2}{c|}{(\FRWJ)}\\
      \hhline{~~~~~~~------}
  \multicolumn{4}{|c|}{}& & & & \textbf{{\scriptsize CBOW}} & \textbf{{\scriptsize SG}} & \textbf{{\scriptsize DM}} & \textbf{{\scriptsize DBOW}}  & \textbf{{\scriptsize DM}} & \textbf{{\scriptsize DBOW}} \\
      \hline
\multicolumn{3}{|c|}{\multirow{3}{*}{\rotatebox[origin=c]{90}{ {\tiny Hierarchical}}}}   &    \textbf{{\tiny WP}} & \Number{1.0000} & \Gray & \Gray & \Gray & \Gray & \Gray & \Gray & \Gray & \Gray\\
      \hhline{~~~----------}
\multicolumn{3}{|c|}{}  & \textbf{{\tiny JC}} & \Number{0.6346} & \Number{1.0000}  & \Gray & \Gray & \Gray & \Gray & \Gray & \Gray & \Gray\\
      \hhline{~~~----------}
\multicolumn{3}{|c|}{}  & \textbf{{\tiny P}} & \Number{0.0097} & \Number{0.0624} & \Number{1.0000}  & \Gray & \Gray & \Gray & \Gray & \Gray & \Gray\\
      \hline
\multirow{6}{*}{\rotatebox[origin=c]{90}{ Embedding}} & \multirow{2}{*}{\rotatebox[origin=c]{90}{\tiny Word2Vec}} & \multirow{2}{*}{\rotatebox[origin=c]{90}{\tiny (\FRWJ)}} &     \textbf{{\tiny CBOW}} & \Number{0.0581}	& \Number{0.0212}	& \Number{0.0013} & \Number{1.0000}  & \Gray & \Gray & \Gray & \Gray & \Gray\\
      \hhline{~~~----------}
&  & &  \textbf{{\tiny SG}} &  \Number{0.0553} &	\Number{0.0188} &	\Number{-0.0043} &	\Number{0.9412} & \Number{1.0000}  &  \Gray & \Gray & \Gray & \Gray\\
      \hhline{~------------}
  & \multirow{4}{*}{\rotatebox[origin=c]{90}{\tiny Doc2Vec}} & \multirow{2}{*}{\rotatebox[origin=c]{90}{\tiny (\FRWFJ)}} & \textbf{{\tiny DM}} &     \Number{0.0040} & 
                        \Number{-0.0298} &
                        \Number{0.0548} &
                \Number{-0.0626} &
                 \Number{-0.0791} &
                 \Number{1.0000} & \Gray
                  & \Gray
                   & \Gray
                   \\
   
      \hhline{~~~----------}
   & & & \textbf{{\tiny DBOW}} &   \Number{0.0466} &
          \Number{0.0155} &
          \Number{0.0359} &
          \Number{0.0362} &
          \Number{0.0222} &
          \Number{0.5691} &
          \Number{1.0000} & \Gray
           & \Gray
          \\
      \hhline{~~-----------}
   & & \multirow{2}{*}{\rotatebox[origin=c]{90}{\tiny (\FRWJ)}} &  \textbf{{\tiny DM}} &     \Number{0.0259} &
           \Number{0.0186} &
           \Number{0.0175} &
           \Number{-0.1865} &
           \Number{-0.2593} &
           \Number{0.1724} &
           \Number{0.1484} & 
           \Number{1.0000} & \Gray
           \\
      \hhline{~~~----------}
   & & & \textbf{{\tiny DBOW}} &   \Number{0.0361} &
        \Number{0.0287} &
        \Number{0.0453} &
        \Number{-0.0896} &
        \Number{-0.1601} &
        \Number{0.1511} &
        \Number{0.1825} & 
        \Number{0.5493} & 
        \Number{1.0000} \\
      \hline

\end{tabular}
\caption{Pearson Correlation for the pairwise text similarities across multiple similarity metrics: 1) WP -- Wu \& Palmer similarity 2) JC -- Jiang-Conrath Similarity 3) P -- Poincaré metric 4) CBOW -- Continuous Bag of Words 5) SG -- Skip-gram 6) DM -- Distributed Memory 7) DBOW -- Distributed Bag of Words}
\label{tab:correlation}
\end{center}
\end{table*}

We have briefly mentioned at the start of this section, we specifically used the first paragraph only text to assert for its goodness compared to the whole text. The rationalé for this as follows: as discussed in subsection \ref{subsec:first_link}, the first paragraph or the lead section of a wiki article is an abstractive summary of the entire article. Hence, if this showed comparable results to full text, the full text document embedding can be substituted by this. Which requires less computational resources due to its much smaller size. For comparison, the word count on data sets \FRWJ{} and \FRWFJ{} are 9,189,536 and 980,047 respectively, which is a 10:1 compression ratio. Further, this would open the door to future in-domain text summarizing research.


\section{Use Case Analysis 2: \FRWTGen}
Our work on free text generation is divided into two phases. In the first phase, we try to train a GPT-Neo~\cite{gao2020pile} model capable of generating text in the Dungeons \& Dragons (Forgotten Realms) domain. Once the text generator was in place, the second phase of the work involves classifying in-domain named entities present in the generated text. 
We used the trained model from step 1 and fine-tuned it using data from the Forgotten Realms Wiki JSON (\textit{FRW-J}) database. 
We tokenized each of the the documents in the \textit{FRW-J} data set separately. 
The model that was used contained 1.3 billion parameters. Using this fine-tuned model, we were able to generate texts that are grounded in the D\&D domain. 

\subsection{Generated Text}
An example text generated using the \FRWTGen{} we have trained using the \FRW{} is shown in Figure \ref{fig:sample_text}. To make comparisons as to the effect of training on the \FRW{} (fine-tuning, we have also listed some text samples generated using the unmodified (vanilla) version of the same language model (Figure~\ref{fig:sample_text_vanilla}). All of these samples were generated based on the same input prompt ``\textit{A Bard, a Wizard, and a Paladin walk into a tavern}''. 

\begin{figure}[!htb]
    \begin{framed}
        \scriptsize
        \textbf{A Bard, a Wizard, and a Paladin walk into a tavern}, and debrief the group on their travels.  The trio discuss their lives and lives. Bard: ``In actuality, the only things I have gained in my life that I truly value are knowledge and comradeship, both of which seem more valuable than wealth.'' Wizard: ``Clan Bloodforge was once my clan; I guess I'm considering it my clan after death to boot.''. Paladin: ``There are two separate communities of people living in this land, one in the North and another in the South, and the northern people came to the west to the Sword Coast about five centuries ago.''  One of Bloodforge's leaders meets with the leaders of Clan K'yorlhan, the tribe from Tethyr that fights the bard raider bandit leader Korgh, and discovers that most of the members of Clan Bloodforge do not want to join Clan K'yorlhan. In the end, he leads them to the hidden Bloodforge, and joins them in battle.  Monk: ``Korgh is a madman, but he's got a point about the invaders.''. Bard: ``That's true; otherwise, why'd he attack all across Tethyr? That's something Clan Bloodforge are going to have to consider.'' During the homecoming party, Bard, Wizard, and Paladin trade news of their adventures with each other. Bard: ``Clan Bloodforge and Clan K'yorlhan, now that's something I'm going to have to watch carefully.''. Wizard: ``I'd better keep in mind that Clan K'yorlhan is already taken.''.  Paladin: ``We were also allied with the tanarukks of Norland, but I hope I'm wrong, because they keep countless spies hidden around here.''.  Paladin - Son of Blackstaff - joins Bloodforge as a young man. Bard: ``Artemis Entreri is not the most forward-thinking of consorts, but still, he's a good man.''  Paladin - A young warrior from the north. Bard: ``Welcome to Bloodforge.''  Paladin - A young man from the west. Bard: ``The Wyrmskull Thron is a noble house in Calimshan.''. Wizard: ``I'd have to say that the War Wizard of Calimshan is a good man.''
        
        \end{framed}
    \caption{Sample text generated from the \FRWTGen{} (trained on \FRW)}
    \label{fig:sample_text}
\end{figure}

\begin{figure}[!htb]
    \centering
    \begin{subfigure}[b]{1\columnwidth}
        \begin{framed}
        \scriptsize
        \textbf{A Bard, a Wizard, and a Paladin walk into a tavern}, It was the middle of the night, and - like the rest of the town - Bard, Wizard, and Paladin had already found a place that they could call their own. It was a place where they could help young minds and provide a positive role model for young people. "Do you mind if I sit?" a BaldDwarf asked. "The kids love your games," the Bard replied. The Bard was an active member of the Order of the Phoenix, the Order of the Emerald Tablet, an Amity Group, the Viscount of Ce's Guild, and anointed as a Paladin by a bard named D'Cel. He was also a Wilderness Shield Master and a beardless man. He couldn't wear any clothing he didn't own, because - to him - the Wilds were his home. The Wizard and Paladin should never have come into a tavern, but the Bard was a musketeer, and a Tourney-member. He was just trying to get a look at the other two. 
        \caption{Sample 1}
        \label{subfig:sample_text_vanilla_1}
        \end{framed}
    \end{subfigure}
    \begin{subfigure}[b]{1\columnwidth}
        \begin{framed}
        \scriptsize
        \textbf{A Bard, a Wizard, and a Paladin walk into a tavern}, in rural Northumberland.  The Bard and the Wizard pay and are received in the company of a Salmon, and the Paladin is greeted as a traveling merchant by merchants and peasants alike. The Bard is either still in his role as an entertainer, or has passed into the service of some people who know what they are doing. The Wizard has given up his peddling.  He is now performing services as an entertainer, singing and dancing and acting as peacemaker or mediator. The Paladin has come here to seek the largess of the Bard, to borrow from him with interest, and to bestow upon him the reward of his protection and the services of his wizardry. The Bard has accepted his terms, and has given his lodgings - a mansion built in a ruined mill, called Cinann House, on the banks of the River Barrow, once belonging to the Earl of Northumberland, who was the brother of stood as guardian of the booth of Alderney who came to trade with Scandinavia to the east.
        \caption{Sample 2}
        \label{subfig:sample_text_vanilla_2}
        \end{framed}
    \end{subfigure}
    \caption{Sample text generated from the vanilla text generator}
    \label{fig:sample_text_vanilla}
\end{figure}

We can observe, that the vanilla model refers to real world entities such as \textit{Northumberland}, \textit{Emerald Tablet}, and \textit{Alderney}. It is also worth noting that the vanilla model refers to other works of literature as can be observed in Figure \ref{subfig:sample_text_vanilla_1} where \textit{Order of the Phoenix\footURL{https://harrypotter.fandom.com/wiki/Order_of_the_Phoenix}} was mentioned. Here, the entity was correctly identified as a group or a society in accordance with its literature and has been used appropriately in the generated text.

In comparison, the text samples generated from the \FRWTGen{} show more \dnd{} domain specific characteristics. It uses established entities from the \dnd{} lore such as 
\textit{Bloodforge}, \textit{Norland}, and \textit{Calimshan}. It also identifies and uses \textit{Norland} as a location which is part of \textit{Sword Coast} in accordance with the domain data. Another thing to note is that the model even generates fake names and characters that are not mentioned in the data set such as \textit{Korgh} and \textit{K'yorlhan} that fit in well with the fantasy genre and build narratives around those characters. Despite, some minor issues with cohesion, overall, it generates satisfactory results. 


\subsection{Named entity classifier}

Although the \FRWTGen{} managed to create text based on \dnd{} domain, when observed carefully by domain experts, there were some inconsistencies with the established lore of the domain. For example, according to the Forgotten Realms lore, \textit{Artemis Entreri} is an \textit{assassin} and not a \textit{consort} while the \textit{Wyrmskull Throne} is a physical object, not the name of a house. To assess the categorical validity of the named entities generated in the text, we have trained the same model on a data set where each row contains a full text generated by the \FRWTGen{}, a named entity in that text, and the matching category extracted from the \FRWI. By performing 5-fold cross validation, we were able to train our model to identify the category for a named entity in a generated text. 
For this basic analysis, we created 100 instances each containing on average 351.4 words and 19.07 sentences. The model was capable of predicting the correct category with 99.3\% accuracy on average, attesting to the power of GPT-Neo~\cite{gpt-neo} as well as the potential in domain specific text generation.
Since the correct classifications are a set of rules declared by the \FRWI{} data set, and the GPT-Neo model uses a data driven training approach, this can be the first step towards creating a conditional text generator that will bridge the traditional rule-based text generation methods and the novel data-driven methods. 

As for the vanilla model, we were unable to perform any meaningful entity classification in relation to the \dnd{} domain, as there were no \dnd{} specific entities that were mentioned in the generated text.



\section{Conclusion and Future Work}

When performing domain specific text generation, it is important that the output stays true to source material. For this, sufficient data from the domain is required. Other than the raw corpora, additional supplementary data structures such as tabular summaries can help ease the process of evaluating the consistency of generated text in context to the domain. In this paper we present a data set based on the \dnd{} domain and a system that is capable of generating free text that stays consistent to the domain. Apart from this, the named entity classifier model shows promising results as part of a guided text generation system.
We hope that the \FR{} offers a convenient unique data set for the \dnd{} domain. We hope that the data set can also be enhanced in the future including an improved linked list to measure evaluation.
%

\bibliography{anthology,custom}
\bibliographystyle{acl_natbib}




\end{document}

%% file: footnotes.tex
\newcommand{\footURL}[1]{\footnote{\url{#1}}}

\makeatletter
\newcommand\footnoteref[1]{\protected@xdef\@thefnmark{\ref{#1}}\@footnotemark}
\makeatother

%% file: acl_latex.bbl
\begin{thebibliography}{57}
\expandafter\ifx\csname natexlab\endcsname\relax\def\natexlab#1{#1}\fi

\bibitem[{Amin-Nejad et~al.(2020)Amin-Nejad, Ive, and
  Velupillai}]{amin-nejad-etal-2020-exploring}
Ali Amin-Nejad, Julia Ive, and Sumithra Velupillai. 2020.
\newblock \href {https://aclanthology.org/2020.lrec-1.578} {Exploring
  transformer text generation for medical dataset augmentation}.
\newblock In \emph{Proceedings of the 12th Language Resources and Evaluation
  Conference}, pages 4699--4708, Marseille, France. European Language Resources
  Association.

\bibitem[{Barzilay and Lapata(2005)}]{barzilay-lapata-2005-collective}
Regina Barzilay and Mirella Lapata. 2005.
\newblock \href {https://aclanthology.org/H05-1042} {Collective content
  selection for concept-to-text generation}.
\newblock In \emph{Proceedings of Human Language Technology Conference and
  Conference on Empirical Methods in Natural Language Processing}, pages
  331--338, Vancouver, British Columbia, Canada. Association for Computational
  Linguistics.

\bibitem[{Black et~al.(2021)Black, Gao, Wang, Leahy, and Biderman}]{gpt-neo}
Sid Black, Leo Gao, Phil Wang, Connor Leahy, and Stella Biderman. 2021.
\newblock \href {https://doi.org/10.5281/zenodo.5297715} {{GPT-Neo: Large Scale
  Autoregressive Language Modeling with Mesh-Tensorflow}}.

\bibitem[{Black et~al.(2022)Black, Biderman, Hallahan, Anthony, Gao, Golding,
  He, Leahy, McDonell, Phang, Pieler, Prashanth, Purohit, Reynolds, Tow, Wang,
  and Weinbach}]{black-etal-2022-gpt}
Sidney Black, Stella Biderman, Eric Hallahan, Quentin Anthony, Leo Gao,
  Laurence Golding, Horace He, Connor Leahy, Kyle McDonell, Jason Phang,
  Michael Pieler, Usvsn~Sai Prashanth, Shivanshu Purohit, Laria Reynolds,
  Jonathan Tow, Ben Wang, and Samuel Weinbach. 2022.
\newblock \href {https://doi.org/10.18653/v1/2022.bigscience-1.9}
  {{GPT}-{N}eo{X}-20{B}: An open-source autoregressive language model}.
\newblock In \emph{Proceedings of BigScience Episode {\#}5 -- Workshop on
  Challenges {\&} Perspectives in Creating Large Language Models}, pages
  95--136, virtual+Dublin. Association for Computational Linguistics.

\bibitem[{Brown et~al.(2020)Brown, Mann, Ryder, Subbiah, Kaplan, Dhariwal,
  Neelakantan, Shyam, Sastry, Askell et~al.}]{brown2020language}
Tom Brown, Benjamin Mann, Nick Ryder, Melanie Subbiah, Jared~D Kaplan, Prafulla
  Dhariwal, Arvind Neelakantan, Pranav Shyam, Girish Sastry, Amanda Askell,
  et~al. 2020.
\newblock Language models are few-shot learners.
\newblock \emph{Advances in neural information processing systems},
  33:1877--1901.

\bibitem[{Busemann and Horacek(1998)}]{busemann-horacek-1998-flexible}
Stephan Busemann and Helmut Horacek. 1998.
\newblock \href {https://aclanthology.org/W98-1425} {A flexible shallow
  approach to text generation}.
\newblock In \emph{Natural Language Generation}, Niagara-on-the-Lake, Ontario,
  Canada. Association for Computational Linguistics.

\bibitem[{Chen et~al.(2021)Chen, Takamura, and
  Nakayama}]{chen-etal-2021-scixgen-scientific}
Hong Chen, Hiroya Takamura, and Hideki Nakayama. 2021.
\newblock \href {https://doi.org/10.18653/v1/2021.findings-emnlp.128}
  {{S}ci{XG}en: A scientific paper dataset for context-aware text generation}.
\newblock In \emph{Findings of the Association for Computational Linguistics:
  EMNLP 2021}, pages 1483--1492, Punta Cana, Dominican Republic. Association
  for Computational Linguistics.

\bibitem[{Cozza et~al.(2016)Cozza, Petrocchi, and Spognardi}]{cozza2016matter}
Vittoria Cozza, Marinella Petrocchi, and Angelo Spognardi. 2016.
\newblock \href {https://doi.org/10.1007/978-3-319-38791-8_31} {{A matter of
  words: NLP for quality evaluation of Wikipedia medical articles}}.
\newblock In \emph{International Conference on Web Engineering}, pages
  448--456. Springer.

\bibitem[{Crawford et~al.(2014)Crawford, Wyatt, Schwalb, and
  Cordell}]{crawford2014player}
Jeremy Crawford, James Wyatt, Robert~J Schwalb, and Bruce~R Cordell. 2014.
\newblock \emph{Player's handbook}.
\newblock Wizards of the Coast LLC.

\bibitem[{Dale et~al.(2003)Dale, Geldof, and Prost}]{dale2003coral}
Robert Dale, Sabine Geldof, and Jean-Philippe Prost. 2003.
\newblock Coral: Using natural language generation for navigational assistance.
\newblock In \emph{Proceedings of the 26th Australasian computer science
  conference-Volume 16}, pages 35--44.

\bibitem[{de~Silva and Dou(2021)}]{de-silva-dou-2021-semantic}
Nisansa de~Silva and Dejing Dou. 2021.
\newblock \href {https://doi.org/10.18653/v1/2021.eacl-main.31} {Semantic
  oppositeness assisted deep contextual modeling for automatic rumor detection
  in social networks}.
\newblock In \emph{Proceedings of the 16th Conference of the European Chapter
  of the Association for Computational Linguistics: Main Volume}, pages
  405--415, Online. Association for Computational Linguistics.

\bibitem[{de~Silva et~al.(2017)de~Silva, Dou, and Huang}]{de2017discovering}
Nisansa de~Silva, Dejing Dou, and Jingshan Huang. 2017.
\newblock \href {https://doi.org/10.1145/3107411.3107452} {{Discovering
  Inconsistencies in PubMed Abstracts Through Ontology-Based Information
  Extraction}}.
\newblock In \emph{Proceedings of the 8th ACM International Conference on
  Bioinformatics, Computational Biology, and Health Informatics}, pages
  362--371.

\bibitem[{El-Kassas et~al.(2020)El-Kassas, Salama, Rafea, and
  Mohamed}]{el2020automatic}
Wafaa~S El-Kassas, Cherif~R Salama, Ahmed~A Rafea, and Hoda~K Mohamed. 2020.
\newblock \href {https://doi.org/10.1016/j.eswa.2020.113679} {Automatic text
  summarization: A comprehensive survey}.
\newblock \emph{Expert Systems with Applications}, page 113679.

\bibitem[{Ferrari et~al.(2017)Ferrari, Donati, and
  Gnesi}]{ferrari2017detecting}
Alessio Ferrari, Beatrice Donati, and Stefania Gnesi. 2017.
\newblock Detecting domain-specific ambiguities: an nlp approach based on
  wikipedia crawling and word embeddings.
\newblock In \emph{2017 IEEE 25th International Requirements Engineering
  Conference Workshops (REW)}, pages 393--399. IEEE.

\bibitem[{Ferschke(2014)}]{ferschke2014quality}
Oliver Ferschke. 2014.
\newblock The quality of content in open online collaboration platforms:
  Approaches to nlp-supported information quality management in wikipedia.

\bibitem[{Ferschke et~al.(2013)Ferschke, Daxenberger, and
  Gurevych}]{ferschke2013survey}
Oliver Ferschke, Johannes Daxenberger, and Iryna Gurevych. 2013.
\newblock A survey of nlp methods and resources for analyzing the collaborative
  writing process in wikipedia.
\newblock In \emph{The People’s Web Meets NLP}, pages 121--160. Springer.

\bibitem[{Fu et~al.(2018)Fu, Tan, Peng, Zhao, and Yan}]{fu2018style}
Zhenxin Fu, Xiaoye Tan, Nanyun Peng, Dongyan Zhao, and Rui Yan. 2018.
\newblock \href {https://doi.org/10.1609/aaai.v32i1.11330} {Style transfer in
  text: Exploration and evaluation}.
\newblock In \emph{Proceedings of the AAAI Conference on Artificial
  Intelligence}, volume~32.

\bibitem[{Gao et~al.(2020)Gao, Biderman, Black, Golding, Hoppe, Foster, Phang,
  He, Thite, Nabeshima et~al.}]{gao2020pile}
Leo Gao, Stella Biderman, Sid Black, Laurence Golding, Travis Hoppe, Charles
  Foster, Jason Phang, Horace He, Anish Thite, Noa Nabeshima, et~al. 2020.
\newblock The pile: An 800gb dataset of diverse text for language modeling.
\newblock \emph{arXiv preprint arXiv:2101.00027}.

\bibitem[{Gholipour~Ghalandari et~al.(2020)Gholipour~Ghalandari, Hokamp, Pham,
  Glover, and Ifrim}]{gholipour-ghalandari-etal-2020-large}
Demian Gholipour~Ghalandari, Chris Hokamp, Nghia~The Pham, John Glover, and
  Georgiana Ifrim. 2020.
\newblock \href {https://doi.org/10.18653/v1/2020.acl-main.120} {A large-scale
  multi-document summarization dataset from the {W}ikipedia current events
  portal}.
\newblock In \emph{Proceedings of the 58th Annual Meeting of the Association
  for Computational Linguistics}, pages 1302--1308, Online. Association for
  Computational Linguistics.

\bibitem[{Gygax and Arneson(1974)}]{gygax1974dungeons}
Gary Gygax and Dave Arneson. 1974.
\newblock \emph{dungeons \& dragons}, volume~19.
\newblock Tactical Studies Rules Lake Geneva, WI.

\bibitem[{Hoffmann et~al.(2010)Hoffmann, Zhang, and
  Weld}]{hoffmann-etal-2010-learning}
Raphael Hoffmann, Congle Zhang, and Daniel~S. Weld. 2010.
\newblock \href {https://aclanthology.org/P10-1030} {Learning 5000 relational
  extractors}.
\newblock In \emph{Proceedings of the 48th Annual Meeting of the Association
  for Computational Linguistics}, pages 286--295, Uppsala, Sweden. Association
  for Computational Linguistics.

\bibitem[{Jiang and Conrath(1997)}]{jiang-conrath-1997-semantic}
Jay~J. Jiang and David~W. Conrath. 1997.
\newblock \href {https://aclanthology.org/O97-1002} {Semantic similarity based
  on corpus statistics and lexical taxonomy}.
\newblock In \emph{Proceedings of the 10th Research on Computational
  Linguistics International Conference}, pages 19--33, Taipei, Taiwan. The
  Association for Computational Linguistics and Chinese Language Processing
  (ACLCLP).

\bibitem[{Kreutzer et~al.(2022)Kreutzer, Caswell, Wang, Wahab, van Esch,
  Ulzii-Orshikh, Tapo, Subramani, Sokolov, Sikasote, Setyawan, Sarin, Samb,
  Sagot, Rivera, Rios, Papadimitriou, Osei, Suarez, Orife, Ogueji, Rubungo,
  Nguyen, M{\"u}ller, M{\"u}ller, Muhammad, Muhammad, Mnyakeni, Mirzakhalov,
  Matangira, Leong, Lawson, Kudugunta, Jernite, Jenny, Firat, Dossou, Dlamini,
  de~Silva, {\c{C}}abuk~Ball{\i}, Biderman, Battisti, Baruwa, Bapna, Baljekar,
  Azime, Awokoya, Ataman, Ahia, Ahia, Agrawal, and
  Adeyemi}]{kreutzer-etal-2022-quality}
Julia Kreutzer, Isaac Caswell, Lisa Wang, Ahsan Wahab, Daan van Esch,
  Nasanbayar Ulzii-Orshikh, Allahsera Tapo, Nishant Subramani, Artem Sokolov,
  Claytone Sikasote, Monang Setyawan, Supheakmungkol Sarin, Sokhar Samb,
  Beno{\^\i}t Sagot, Clara Rivera, Annette Rios, Isabel Papadimitriou, Salomey
  Osei, Pedro~Ortiz Suarez, Iroro Orife, Kelechi Ogueji, Andre~Niyongabo
  Rubungo, Toan~Q. Nguyen, Mathias M{\"u}ller, Andr{\'e} M{\"u}ller,
  Shamsuddeen~Hassan Muhammad, Nanda Muhammad, Ayanda Mnyakeni, Jamshidbek
  Mirzakhalov, Tapiwanashe Matangira, Colin Leong, Nze Lawson, Sneha Kudugunta,
  Yacine Jernite, Mathias Jenny, Orhan Firat, Bonaventure F.~P. Dossou, Sakhile
  Dlamini, Nisansa de~Silva, Sakine {\c{C}}abuk~Ball{\i}, Stella Biderman,
  Alessia Battisti, Ahmed Baruwa, Ankur Bapna, Pallavi Baljekar, Israel~Abebe
  Azime, Ayodele Awokoya, Duygu Ataman, Orevaoghene Ahia, Oghenefego Ahia,
  Sweta Agrawal, and Mofetoluwa Adeyemi. 2022.
\newblock \href {https://doi.org/10.1162/tacl_a_00447} {Quality at a glance: An
  audit of web-crawled multilingual datasets}.
\newblock \emph{Transactions of the Association for Computational Linguistics},
  10:50--72.

\bibitem[{Lamprecht et~al.(2016)Lamprecht, Dimitrov, Helic, and
  Strohmaier}]{lamprecht2016evaluating}
Daniel Lamprecht, Dimitar Dimitrov, Denis Helic, and Markus Strohmaier. 2016.
\newblock Evaluating and improving navigability of wikipedia: a comparative
  study of eight language editions.
\newblock In \emph{Proceedings of the 12th international symposium on open
  collaboration}, pages 1--10.

\bibitem[{Lange et~al.(2010)Lange, B{\"o}hm, and Naumann}]{lange2010extracting}
Dustin Lange, Christoph B{\"o}hm, and Felix Naumann. 2010.
\newblock Extracting structured information from wikipedia articles to populate
  infoboxes.
\newblock In \emph{Proceedings of the 19th ACM international conference on
  Information and knowledge management}, pages 1661--1664.

\bibitem[{Le and Mikolov(2014)}]{le2014distributed}
Quoc Le and Tomas Mikolov. 2014.
\newblock Distributed representations of sentences and documents.
\newblock In \emph{International conference on machine learning}, pages
  1188--1196. PMLR.

\bibitem[{Lebret et~al.(2016)Lebret, Grangier, and
  Auli}]{lebret-etal-2016-neural}
R{\'e}mi Lebret, David Grangier, and Michael Auli. 2016.
\newblock \href {https://doi.org/10.18653/v1/D16-1128} {Neural text generation
  from structured data with application to the biography domain}.
\newblock In \emph{Proceedings of the 2016 Conference on Empirical Methods in
  Natural Language Processing}, pages 1203--1213, Austin, Texas. Association
  for Computational Linguistics.

\bibitem[{Liang et~al.(2009)Liang, Jordan, and
  Klein}]{liang-etal-2009-learning}
Percy Liang, Michael Jordan, and Dan Klein. 2009.
\newblock \href {https://aclanthology.org/P09-1011} {Learning semantic
  correspondences with less supervision}.
\newblock In \emph{Proceedings of the Joint Conference of the 47th Annual
  Meeting of the {ACL} and the 4th International Joint Conference on Natural
  Language Processing of the {AFNLP}}, pages 91--99, Suntec, Singapore.
  Association for Computational Linguistics.

\bibitem[{Liu et~al.(2018)Liu, Wang, Sha, Chang, and Sui}]{liu2018table}
Tianyu Liu, Kexiang Wang, Lei Sha, Baobao Chang, and Zhifang Sui. 2018.
\newblock Table-to-text generation by structure-aware seq2seq learning.
\newblock In \emph{Thirty-Second AAAI Conference on Artificial Intelligence}.

\bibitem[{McRoy et~al.(2003)McRoy, Channarukul, and Ali}]{mcroy2003augmented}
Susan~W McRoy, Songsak Channarukul, and Syed~S Ali. 2003.
\newblock An augmented template-based approach to text realization.
\newblock \emph{Natural Language Engineering}, 9(4):381--420.

\bibitem[{Mikolov et~al.(2013)Mikolov, Chen, Corrado, and
  Dean}]{mikolov2013efficient}
Tomas Mikolov, Kai Chen, Greg Corrado, and Jeffrey Dean. 2013.
\newblock Efficient estimation of word representations in vector space.
\newblock \emph{arXiv preprint arXiv:1301.3781}.

\bibitem[{Nastase and Strube(2013)}]{nastase2013transforming}
Vivi Nastase and Michael Strube. 2013.
\newblock Transforming wikipedia into a large scale multilingual concept
  network.
\newblock \emph{Artificial Intelligence}, 194:62--85.

\bibitem[{Nickel and Kiela(2017)}]{nickel2017poincare}
Maximilian Nickel and Douwe Kiela. 2017.
\newblock \href
  {http://papers.nips.cc/paper/7213-poincare-embeddings-for-learning-hierarchical-representations.pdf}
  {Poincar\'{e} embeddings for learning hierarchical representations}.
\newblock In I.~Guyon, U.~V. Luxburg, S.~Bengio, H.~Wallach, R.~Fergus,
  S.~Vishwanathan, and R.~Garnett, editors, \emph{Advances in Neural
  Information Processing Systems 30}, pages 6341--6350. Curran Associates, Inc.

\bibitem[{Ponzetto and Strube(2007)}]{ponzetto2007knowledge}
Simone~Paolo Ponzetto and Michael Strube. 2007.
\newblock Knowledge derived from wikipedia for computing semantic relatedness.
\newblock \emph{Journal of Artificial Intelligence Research}, 30:181--212.

\bibitem[{Radford et~al.(2019)Radford, Wu, Child, Luan, Amodei, Sutskever
  et~al.}]{radford2019language}
Alec Radford, Jeffrey Wu, Rewon Child, David Luan, Dario Amodei, Ilya
  Sutskever, et~al. 2019.
\newblock Language models are unsupervised multitask learners.
\newblock \emph{OpenAI blog}, 1(8):9.

\bibitem[{Rajapaksha et~al.(2020)Rajapaksha, Mudalige
  et~al.}]{rajapaksha2020rule}
Isanka Rajapaksha, Chanika~Ruchini Mudalige, et~al. 2020.
\newblock \href {https://doi.org/10.1109/ICTer51097.2020.9325435} {{Rule-Based
  Approach for Party-Based Sentiment Analysis in Legal Opinion Texts}}.
\newblock In \emph{2020 20th International Conference on Advances in ICT for
  Emerging Regions (ICTer)}, pages 284--285. IEEE.

\bibitem[{Rajpurkar et~al.(2016)Rajpurkar, Zhang, Lopyrev, and
  Liang}]{rajpurkar-etal-2016-squad}
Pranav Rajpurkar, Jian Zhang, Konstantin Lopyrev, and Percy Liang. 2016.
\newblock \href {https://doi.org/10.18653/v1/D16-1264} {{SQ}u{AD}: 100,000+
  questions for machine comprehension of text}.
\newblock In \emph{Proceedings of the 2016 Conference on Empirical Methods in
  Natural Language Processing}, pages 2383--2392, Austin, Texas. Association
  for Computational Linguistics.

\bibitem[{Rameshkumar and Bailey(2020)}]{rameshkumar-bailey-2020-storytelling}
Revanth Rameshkumar and Peter Bailey. 2020.
\newblock \href {https://doi.org/10.18653/v1/2020.acl-main.459} {Storytelling
  with dialogue: {A} {Critical Role Dungeons and Dragons Dataset}}.
\newblock In \emph{Proceedings of the 58th Annual Meeting of the Association
  for Computational Linguistics}, pages 5121--5134, Online. Association for
  Computational Linguistics.

\bibitem[{Reiter and Dale(1997)}]{reiter1997building}
Ehud Reiter and Robert Dale. 1997.
\newblock Building applied natural language generation systems.
\newblock \emph{Natural Language Engineering}, 3(1):57--87.

\bibitem[{Reiter et~al.(2005)Reiter, Sripada, Hunter, Yu, and
  Davy}]{reiter2005choosing}
Ehud Reiter, Somayajulu Sripada, Jim Hunter, Jin Yu, and Ian Davy. 2005.
\newblock Choosing words in computer-generated weather forecasts.
\newblock \emph{Artificial Intelligence}, 167(1-2):137--169.

\bibitem[{Sanchez-Perez et~al.(2014)Sanchez-Perez, Sidorov, and
  Gelbukh}]{sanchez2014winning}
Miguel~A Sanchez-Perez, Grigori Sidorov, and Alexander~F Gelbukh. 2014.
\newblock A winning approach to text alignment for text reuse detection at pan
  2014.
\newblock In \emph{CLEF (Working Notes)}, pages 1004--1011.

\bibitem[{Srinivasan et~al.(2021)Srinivasan, Raman, Chen, Bendersky, and
  Najork}]{srinivasan2021wit}
Krishna Srinivasan, Karthik Raman, Jiecao Chen, Michael Bendersky, and Marc
  Najork. 2021.
\newblock Wit: Wikipedia-based image text dataset for multimodal multilingual
  machine learning.
\newblock In \emph{Proceedings of the 44th International ACM SIGIR Conference
  on Research and Development in Information Retrieval}, pages 2443--2449.

\bibitem[{Stent et~al.(2004)Stent, Prasad, and
  Walker}]{stent-etal-2004-trainable}
Amanda Stent, Rashmi Prasad, and Marilyn Walker. 2004.
\newblock \href {https://doi.org/10.3115/1218955.1218966} {Trainable sentence
  planning for complex information presentations in spoken dialog systems}.
\newblock In \emph{Proceedings of the 42nd Annual Meeting of the Association
  for Computational Linguistics ({ACL}-04)}, pages 79--86, Barcelona, Spain.

\bibitem[{Sugathadasa et~al.(2017)Sugathadasa, Ayesha
  et~al.}]{sugathadasa2017synergistic}
Keet Sugathadasa, Buddhi Ayesha, et~al. 2017.
\newblock \href {https://doi.org/10.1109/ICIINFS.2017.8300343} {{Synergistic
  Union of Word2Vec and Lexicon for Domain Specific Semantic Similarity}}.
\newblock In \emph{2017 IEEE International Conference on Industrial and
  Information Systems (ICIIS)}, pages 1--6. IEEE.

\bibitem[{Sugathadasa et~al.(2018)Sugathadasa, Ayesha
  et~al.}]{sugathadasa2018legal}
Keet Sugathadasa, Buddhi Ayesha, et~al. 2018.
\newblock \href {https://doi.org/10.1007/978-3-030-01177-2_12} {{Legal Document
  Retrieval using Document Vector Embeddings and Deep Learning}}.
\newblock In \emph{Science and information conference}, pages 160--175.
  Springer.

\bibitem[{Turner et~al.(2009)Turner, Sripada, and
  Reiter}]{turner-etal-2009-generating}
Ross Turner, Yaji Sripada, and Ehud Reiter. 2009.
\newblock \href {https://aclanthology.org/W09-0607} {Generating approximate
  geographic descriptions}.
\newblock In \emph{Proceedings of the 12th {E}uropean Workshop on Natural
  Language Generation ({ENLG} 2009)}, pages 42--49, Athens, Greece. Association
  for Computational Linguistics.

\bibitem[{Whitten(2021)}]{whitten_dungeons_2021}
Sarah Whitten. 2021.
\newblock \href
  {https://www.cnbc.com/2021/03/13/dungeons-dragons-had-its-biggest-year-despite-the-coronavirus.html}
  {Dungeons \& {Dragons} had its biggest year ever as {Covid} forced the game
  off tables and onto the web}.

\bibitem[{Wu and Weld(2010)}]{wu-weld-2010-open}
Fei Wu and Daniel~S. Weld. 2010.
\newblock \href {https://aclanthology.org/P10-1013} {Open information
  extraction using {W}ikipedia}.
\newblock In \emph{Proceedings of the 48th Annual Meeting of the Association
  for Computational Linguistics}, pages 118--127, Uppsala, Sweden. Association
  for Computational Linguistics.

\bibitem[{Wu and Palmer(1994)}]{wu-palmer-1994-verb}
Zhibiao Wu and Martha Palmer. 1994.
\newblock \href {https://doi.org/10.3115/981732.981751} {Verb semantics and
  lexical selection}.
\newblock In \emph{32nd Annual Meeting of the Association for Computational
  Linguistics}, pages 133--138, Las Cruces, New Mexico, USA. Association for
  Computational Linguistics.

\bibitem[{Xing and Paul(2017)}]{xing-paul-2017-incorporating}
Linzi Xing and Michael~J. Paul. 2017.
\newblock \href {https://doi.org/10.18653/v1/W17-4406} {Incorporating metadata
  into content-based user embeddings}.
\newblock In \emph{Proceedings of the 3rd Workshop on Noisy User-generated
  Text}, pages 45--49, Copenhagen, Denmark. Association for Computational
  Linguistics.

\bibitem[{Zesch and Gurevych(2007)}]{zesch-gurevych-2007-analysis}
Torsten Zesch and Iryna Gurevych. 2007.
\newblock \href {https://aclanthology.org/W07-0201} {Analysis of the
  {W}ikipedia category graph for {NLP} applications}.
\newblock In \emph{Proceedings of the Second Workshop on {T}ext{G}raphs:
  Graph-Based Algorithms for Natural Language Processing}, pages 1--8,
  Rochester, NY, USA. Association for Computational Linguistics.

\bibitem[{Zesch et~al.(2007)Zesch, Gurevych, and
  M{\"u}hlh{\"a}user}]{zesch2007analyzing}
Torsten Zesch, Iryna Gurevych, and Max M{\"u}hlh{\"a}user. 2007.
\newblock Analyzing and accessing wikipedia as a lexical semantic resource.
\newblock \emph{Data Structures for Linguistic Resources and Applications},
  197205.

\bibitem[{Zesch et~al.(2008)Zesch, M{\"u}ller, and
  Gurevych}]{zesch-etal-2008-extracting}
Torsten Zesch, Christof M{\"u}ller, and Iryna Gurevych. 2008.
\newblock \href
  {http://www.lrec-conf.org/proceedings/lrec2008/pdf/420_paper.pdf} {Extracting
  lexical semantic knowledge from {W}ikipedia and {W}iktionary}.
\newblock In \emph{Proceedings of the Sixth International Conference on
  Language Resources and Evaluation ({LREC}'08)}, Marrakech, Morocco. European
  Language Resources Association (ELRA).

\bibitem[{Zhang et~al.(2022)Zhang, Song, Li, Zhou, and Song}]{zhang2022survey}
Hanqing Zhang, Haolin Song, Shaoyu Li, Ming Zhou, and Dawei Song. 2022.
\newblock A survey of controllable text generation using transformer-based
  pre-trained language models.
\newblock \emph{arXiv preprint arXiv:2201.05337}.

\bibitem[{Zhang et~al.(2019)Zhang, Kishore, Wu, Weinberger, and
  Artzi}]{zhang2019bertscore}
Tianyi Zhang, Varsha Kishore, Felix Wu, Kilian~Q Weinberger, and Yoav Artzi.
  2019.
\newblock {BERTScore: Evaluating Text Generation with BERT}.
\newblock In \emph{International Conference on Learning Representations}.

\bibitem[{Zhang et~al.(2021)Zhang, Jiang, Shang, Cheng, Zhang, Fan, Xiao, and
  Long}]{zhang2021dsgpt}
Xueying Zhang, Yunjiang Jiang, Yue Shang, Zhaomeng Cheng, Chi Zhang, Xiaochuan
  Fan, Yun Xiao, and Bo~Long. 2021.
\newblock \href {https://doi.org/10.1145/3404835.3463037} {{DSGPT:
  Domain-Specific Generative Pre-Training of Transformers for Text Generation
  in E-commerce Title and Review Summarization}}.
\newblock In \emph{Proceedings of the 44th International ACM SIGIR Conference
  on Research and Development in Information Retrieval}, pages 2146--2150.

\bibitem[{Zhou et~al.(2015)Zhou, He, Zhao, and Hu}]{zhou-etal-2015-learning}
Guangyou Zhou, Tingting He, Jun Zhao, and Po~Hu. 2015.
\newblock \href {https://doi.org/10.3115/v1/P15-1025} {Learning continuous word
  embedding with metadata for question retrieval in community question
  answering}.
\newblock In \emph{Proceedings of the 53rd Annual Meeting of the Association
  for Computational Linguistics and the 7th International Joint Conference on
  Natural Language Processing (Volume 1: Long Papers)}, pages 250--259,
  Beijing, China. Association for Computational Linguistics.

\end{thebibliography}
